\documentclass[11pt]{article}

\usepackage[babel=true,kerning=true]{microtype}
\usepackage{multirow}

\usepackage[latin1]{inputenc}

\usepackage{geometry}   
\usepackage{graphicx}  
\usepackage{flafter}  

\usepackage{amsmath,amssymb}  
\usepackage{bm}  
\usepackage{dsfont}
\usepackage{fourier} 
\usepackage[scaled=0.875]{helvet} 

\usepackage{tikz}
\usetikzlibrary{arrows,automata}
\usepackage{pgfkeys}

\usepackage{setspace}
\newcommand{\ben}{\begin{enumerate}}
\newcommand{\een}{\end{enumerate}}

\newcommand{\bc}{\begin{center}}
\newcommand{\ec}{\end{center}}

\newcommand{\bit}{\begin{itemize}}
\newcommand{\eit}{\end{itemize}}

\newcommand{\ds}{\displaystyle}
\newcommand{\beq}{\begin{equation}}
\newcommand{\eeq}{\end{equation}}

\newcommand{\w}{\mathbf{w}}

\newcommand{\wre}{\mathbf{w}^{\rm{r}}}
\newcommand{\wi}{\mathbf{w}^{\rm{in}}}
\newcommand{\wo}{\mathbf{w}^{\rm{out}}}

\newcommand{\va}{\mathbf{a}}
\newcommand{\vb}{\mathbf{b}}

\newcommand{\y}{\mathbf{y}}

\newcommand{\Na}{N_{\rm{a}}}
\newcommand{\Nb}{N_{\rm{b}}}

\newcommand{\Ns}{N_{\rm{s}}}

\newcommand{\N}{\mathds{N}}
\newcommand{\R}{\mathds{R}}

\newcommand{\vs}{\mathbf{s}}

\newif\ifnotes\notestrue

\def\hgr#1{}

\def\hgr#1{}

\title{Empirical Analysis of the Necessary and Sufficient Conditions of the Echo State Property}
\author{Sebasti\'an Basterrech\thanks{This is a version of an accepted paper that will appear in proceeding of the IEEE International Joint Conference on Neural Networks (IJCNN) 2017.}\\
Department of Computer Science, Faculty of Electrical Engineering\\
Czech Technical University, Praha, Czech Republic\\
\texttt{Sebastian.Basterrech@agents.fel.cvut.cz}
}
\date{~}
\begin{document}
\maketitle

%


\vspace{-0.2in}
\begin{abstract}
The Echo State Network (ESN) is a specific recurrent network, which has gained popularity during the last years.
The model has a recurrent network named reservoir, that is fixed during the learning process. The reservoir is used for transforming the input space in a larger space. 
A fundamental property that provokes an impact on the model accuracy is the Echo State Property (ESP).  
There are two main theoretical results related to the ESP.
First, a sufficient condition for the ESP existence that involves the singular values of the reservoir matrix. 
Second, a necessary condition for the ESP. The ESP can be violated according to the spectral radius value of the reservoir matrix.
There is a theoretical gap between these necessary and sufficient conditions.
This article presents an empirical analysis of the accuracy and the projections of reservoirs that satisfy this theoretical gap.
It gives some insights about the generation of the reservoir matrix.
From previous works, it is already known that the optimal accuracy is obtained near to the border of stability control of the dynamics.
Then, according to our empirical results, we can see that this border seems to be closer to the sufficient conditions than to the necessary conditions of the ESP.
\end{abstract}

\section{Introduction}
Recurrent Neural Networks (RNNs) are fascinating tools for modelling time-series.
At the beginning of the 2000s, the Echo State Network (ESN)~\cite{Jaeger01} and Liquid State Machines (LSM)~\cite{Maass99} have been introduced to the Neural Network community.  
They are RNNs with a specific topology and training procedure.
Both models have been developed independently~\cite{JaegerMaass07}.
During the last years many variations of the original ESN and LSM have been introduced. 
Since around 10 years ago, all those methods became known as Reservoir Computing (RC) models. 
Nowadays, the RC models are very popular due to several characteristics, which can be summarised as: robustness, fast computing, understandable, easy to programming, and good accuracy. 
They have achieved good performance in solving well-known benchmark problems.
In particular, they have been successfully applied to solve temporal learning problems~\cite{Jaeger09}.

A fundamental property of the ESN concerning to the network stability is named Echo State Property (ESP).
The ESP guarantees good designs of the topology of the ESN.
In other words, the model is in a suitable state to do good predictions.
The spectral radius and the singular value of the matrix of recurrent connections (named reservoir matrix) are important parameters of the model. Both parameters impact on the ESP.
Actually, under some algebraic condition the ESP is guaranteed (these conditions are associated with the singular value of the reservoir matrix). 
On the other hand, if some algebraic conditions are presented, then the ESP can be violated (these conditions are associated with the spectral radius of the reservoir matrix).
We can see those conditions as the \textit{necessary} and \textit{sufficient} conditions related to the ESP.

\subsection{Goals and Motivations}
The goal of this article is to analyze the ESN accuracy and the reservoir projections for one specific subset of reservoir matrices. 
We focus on the experimental analysis of a ESN model when the reservoir is defined in such a way that we neither confirm the ESP nor deny the ESP existence.
The main motivations of studying those reservoirs are the following ones:
\bit 
\item Unfortunately, there is a theoretical gap between the necessary and sufficient conditions. There are reservoir matrices which we can not affirm if the ESP is holds or not~\cite{Wainrib15}.
\item This gap is big~\cite{Zhang12}, we neither confirm the ESP nor deny the ESP on a large and rich family of reservoir matrices.
\item There are literature that suggests that the optimal computational performance of the reservoir units operates in a regime that lies between stable and chaotic behaviour~\cite{Verstraeten07,Legenstein07,Langton90}. 
\item It seems that the ESN operates optimally in a stable situation when the projections are close to the border of instability~\cite{Jaeger09,Verstraeten07}.
\item The RC models are widely used for solving supervised learning problems. 
The initialization of the reservoir impacts on the model accuracy. As a consequence, an algorithm for generating optimal reservoirs is extremely valuable in the community. Here we are analyzing the best way for scaling the random initialized reservoirs.
\eit

\subsection{Temporal Supervised Learning}
Given a dataset $\mathcal{L}$ with $T$ pairs of inputs $\va(t)\in\mathcal{A}$ of dimension $\Na$ and desired outputs $\vb(t)\in\mathcal{B}$ of dimension $\Nb$, the goal is finding a model $\psi(\w,\cdot)$ such that $\psi(\w,\va(t))$ approximates ``better'' as possible $\vb(t)$ for all $\va(t)$ in $\mathcal{L}$. 
We denote by $\w$  the undefined parameters of the model, which are adjusted according to the dataset $\mathcal{L}$.
Let $\y(\va(t))$ be the output vector produced by the model $\psi(\w,\cdot)$ when the input is $\va(t)$.
In order of assessing the accuracy of the model a cost function is defined, which is a distance between the predictions $\y(\w,\va(t))$ and the target $\vb(t)$, here we use the  Normalized Root Mean Squared Error (NRMSE)~\cite{Jaeger09}:
\begin{equation}
\label{NRMSE}
E(\y(t),b(t))=\displaystyle{\sqrt{\frac{\langle||\vb(t) - \y(\w,\va(t)) ||^2 \rangle}{\langle||\vb(t)-\langle \vb(t)\rangle||^2\rangle}}},
\end{equation}
where $||\cdot||$ denotes the Euclidean norm and $\langle\cdot\rangle$ denotes the empirical mean function.
For the sake of the simplicity notation, we denote the model output only according the time index $\y(t)$, instead of $\y(\w,\va(t))$.

There are at least two well-differentiated situations in supervised learning. In one case, each data point of $\mathcal{L}$ is independent of each other. 
This context is called non-temporal supervised learning. On another case, $\mathcal{L}$ contains  dependent data points. This situation is named temporal learning.
Even though, the RNNs and its variations can be used for solving non-temporal learning problems, the most common applications are on the context of temporal learning.
In this case, the model has the form $\psi(\w,\va(t),\va(t-1),\ldots)$ due to the fact  that each point is dependent of each other one.

\section{Reservoir Computing Models}
\label{RC_Models}
\begin{figure}
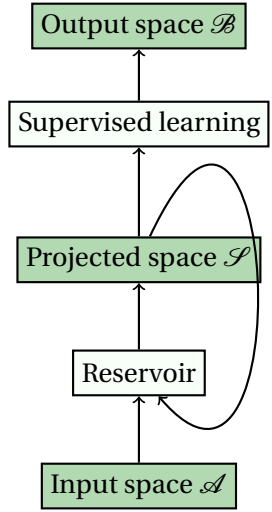

\begin{center}
\tikz{
\node[draw,thick,fill=black!50!green!30, rectangle,minimum size=3.5ex,align=center]  (id0) at (0,-1.5) {Input space $\mathcal{A}$};
      \node[draw,thick,fill=green!3, minimum size=3.5ex]  (id1) at (0,0) {Reservoir};
\node[draw,thick,fill=black!50!green!30, rectangle,minimum size=3.5ex,align=center]  (id2) at (0,1.5) {Projected space $\mathcal{S}$};
\node[draw,thick,fill=green!3, minimum size=3.5ex,align=center]  (id3) at (0,3.3) {Supervised learning};
\node[draw,thick,fill=black!50!green!30, rectangle,minimum size=3.5ex,align=center]  (id4) at (0,4.6) {Output space $\mathcal{B}$};
\draw[->,black] (id0)  edge[->,thick] (id1);
\draw[->,black] (id1)  edge[->,thick] (id2);
\draw[->,black] (id2)  edge[->,thick] (id3);
\draw[->,black] (id3)  edge[->,thick] (id4);
\path[->,black,every loop/.style={looseness=10},distance=4cm,in distance=3cm] (id2) edge[->,thick,in=-50,out=65] (id1); 
}
\end{center}
\caption{The conceptual scheme of a Reservoir Computing model.}
\label{Reservoirschema}  
\end{figure}

The Reservoir Computing (RC) paradigm has started around 15 years ago with the introduction of a new approach for designing the topology and the training algorithms of Recurrent Neural Networks (RNNs).
There is a general consensus that the first two models presented to the community are: Echo State Network (ESN)~\cite{Jaeger01} and Liquid State Machine (LSM)~\cite{Maass99}.
Since 2007, these methods and their variations have started to be popular under the name Reservoir Computing models~\cite{Verstraeten07}.
A RC method has two types of well-distinguished structures. 
One is a RNN which  parameters (weight connections) are random initizialized  and fixed during the learning process. 
Another structure is memory-free (without recurrences) and its parameters are adjusted using traditional approaches of supervised learning.
The memory-free structure is often called readout, and most often consists of a linear regression model.
Figure~\ref{Reservoirschema} shows a general scheme of the information flow of a RC model.
The reservoir structure projects the input patterns in a new larger space. 
This projection has the following two goals: one is to enhance the linear separability of the input space, another one is to memorize the sequence of input patterns.
A linear regression is applied from the projected space to the output space for generating the model outputs.

There are several types of RC models. Although, the main difference among them is the kind of activation function in the reservoir nodes and the type of supervised learning tool in the memory-free structure. For example, the LSM arises  from the interest in making a conceptual representation of the cortical microstructures in the brain, the neurons on the reservoir are LIF neurons~\cite{Maass02}. 
A RC model named Leaky integrator ESN introduced in~\cite{Jaeger07} has gained popularity  due to its well results in practice~\cite{Jaeger07,Jaeger09}. 
In this model each neuron has a weighted memory about its previous state. Then, the variation of the neuron state is much more smooth than in the case of classic sigmoid neurons.
A reservoir with dynamical synapses and threshold logic rates has been studied in~\cite{Verstraeten07}. 
Reservoir units with presence of noise has been studied in~\cite{Rodan11}.  
In addition, two models with neurons inspired from recursive Self-Organizing Maps (SOMs) have been also developed in~\cite{Luko10,BasterCord11}.
Another RC model that combines ideas from another scientific area has been introduced in~\cite{Baster12ESQN}, in this case the activations are based on queueing network behaviour.
The presented list of RC model examples isn't exhaustive. 
All these models have in common that they have a specific type of projection from the input patterns in a large space. These projections have a type of memory given by recurrences on the network, and their parameters remain fixed during the training.
%

%


%
%
\newcommand{\Nu}{N_{\rm{u}}}
\newcommand{\vu}{\mathbf{u}}
%

%
\subsection{Mathematical Formalization of the Echo State Network Model}
We are following the previous notation. Given a learning dataset $\mathcal{L}$ with inputs $\va(t)\in\mathcal{A}$ of dimension $\Na$, a reservoir is a RNN composed by $\Ns$ interconnected neurons. 
The connections are collected in $\Ns\times\Ns$ matrix that we denote by $\wre$.
A matrix $\wi$ with dimensions $\Na\times\Ns$ collects the forward weights between the inputs and reservoir neurons.
For notation simplicity, we include the bias in $\wi$.
We assume discrete dynamics, then at each time step $t$ an input pattern $\va(t)$ is presented to the network, and the reservoir is computed following the recurrent expression
\begin{equation}
\label{reservoirState}
\vs(t+1)=\psi(\wi \va(t+1)+\wre \vs(t)),
\end{equation}
where $\psi(\wi,\wre,\cdot)$ is a Lipschitz function~\cite{Wainrib15}, most often is the hyperbolic tangent function.
We can see the reservoir as an independent RNN from an input space $\mathcal{A}$ to $\mathcal{S}$ that expands the history of input data $(\va(t),\va(t-1),\ldots)$ into a space of dimension $\Ns$. 
The reservoir size is selected such that $\Na\ll \Ns$.
Once the  projections from $\mathcal{A}$ to $\mathcal{S}$ are performed, a parametric function $\nu:\mathcal{S}\rightarrow\mathcal{B}$ is learnt using the training samples in $\L$.
In the canonical ESN the function $\nu(\wo, \cdot)$ is a linear model, and its parameters ($\wo$) are the forwards weights between the reservoir neurons and the output neurons. We collect those weights in a $\Nb\times\Ns$ matrix. Again, we avoid the bias term of the linear regression in $\wo$.
The model output is computed as
\begin{equation}
\label{readout}
\y(t)=\nu(\wo,\vs(t)) = \wo\vs(t).
\end{equation}
A popular training algorithm for computing $\wo$ in the expression~(\ref{readout}) is the offline ridge regression~\cite{Jaeger01}.
The algorithm uses two auxiliary matrices $\mathbf{S}$ and $\mathbf{B}$ of dimensions $\Ns\times T$ and $\Nb\times T$, respectively. 
These matrices collect in their rows the reservoir projections $\vs{(t)}$ and target data $\vb{(t)}$.
Then, the output weight matrix $\wo$ is computed by 
\begin{equation}
\wo=\mathbf{B}\mathbf{S}^T(\mathbf{S}\mathbf{S}^T+\gamma^2\mathbf{I})^{-1},
\end{equation}
where $\mathbf{I}$ is the identity matrix of rank $\Ns$, $\gamma$ is a regularization parameter, and the matrices $\mathbf{B}\mathbf{S}^T$ and $\mathbf{S}\mathbf{S}^T$ have dimensions $\Nb\times\Ns$ and $\Ns\times\Ns$, respectively.
%
%
As a consequence, the solution complexity does not depend on the number of samples, neither in time or in space~\cite{Jaeger09}.

\subsection{Properties of the ESN projections}
The ESN belongs to the family of random projection models~\cite{Butcher2013}.
The model is based on the fact that a random encoding of the input samples can enhance their linear separability.
Even though the trajectories of the reservoir states are random initialized, the model should be independent of the initial network trajectories in the long term.
Then, the network needs to have some type of fading memory with respect of the initial conditions and initial dynamics.
Additionally, it should satisfy a type of ``functional'' relationship where each input sequence has a single output sequence in the long term.
These two characteristics are established in a property regarding the transitions of the reservoir states named Echo State Property (ESP)~\cite{Jaeger09}.
In the following we present the ESP~\cite{Jaeger01}. It is assumed that the network topology hasn't got feedback connections, the input sequences belong to an input space $\mathcal{A}$, and the network states are in a compact set $\mathcal{S}$, then ESN has echo states if $\vs(t)$ is uniquely determined by any left-infinite input sequence $\{\va(t-k):k\in \N\}$~\cite{Zhang12}.
%
%
The ESP establishes  that the trajectories of reservoir states only depend of the input driven network, it doesn't depend on the initial conditions of the network. 
In other words, similar reservoir states must be generated for similar input sequences.
If the model doesn't satisfy the ESP, then it implies that small perturbations can bring the network to new states, which can impact on the prediction abilities of the model~\cite{Wainrib15}.

We specify some notation, let $\rho(A)$ be the spectral radius of a matrix $A$, and let $\eta(A)$ be the singular value of $A$.
The following fundamental result has been analyzed in~\cite{Jaeger01,Jaeger09,Wainrib15}: if the maximum singular value of the reservoir connexion matrix is bounded, then the model satisfy the ESP for every input.
In more detail, if $\eta(\wre)<1$ (which is defined as $\sqrt{\rho(\wre{\wre}^T)}$,  where ${\wre}^T$ is the transposed reservoir matrix) then the ESP is held for every input. 
On the other hand, the ESP is violated when the $\rho(\wre)>1$, with the additional condition that $\mathcal{A}$ contains the zero input sequence.
As a consequence, $\rho(\wre)\leq 1$ is used as a necessary condition for the ESP. 
In addition, the ESP can be lost even for $\rho(\wre)< 1$ (e.g. in zero-input case), and vice-verse, the ESP can be preserved for  $\rho(\wre)>1$~\cite{Manjunath13}.

Therefore, there are two well-analysed situations, a sufficient and a necessary condition related to the ESP. In summary, we have:
\bit
\item \textit{Sufficient condition}: if the $\eta(\wre)<1$, then the ESP is satisfied.
\item \textit{Necessary condition}: it is necessary that $\rho(\wre)\leq 1$ in order of holding the ESP.
\eit

A simple procedure for creating an ESN is to randomly initialize the reservoir matrix $\wre_{\rm{initial}}$ and then to scale it using a factor $\alpha$ as follows: $\wre=\alpha\wre_{\rm{initial}}$.
The selection of the scaling factor impacts on the ESP. 
The sufficient condition to hold the ESP states that $\alpha<\eta(\wre)^{-1}$, and the necessary condition states that $\alpha<\rho(\wre)^{-1}$.
In practice, to use the sufficient condition can be conservative. Furthermore, it can produce a negative impact on the long memory capacity of the reservoir~\cite{Jaeger01,Zhang12}. The sufficient condition can be too restrictive. On the other hand, if is violated the necessary condition ($\rho(\wre)>1$) the network has an asymptotically unstable null state thus, the ESP is lost for any input set containing a zero-input pattern~\cite{Jaeger01}.

\begin{figure}[h]
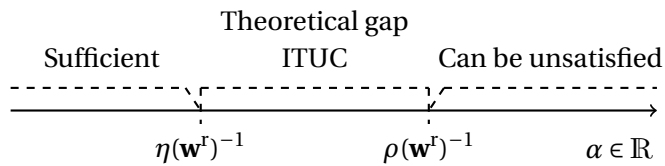

\vspace{0.2in}
\begin{center}
	\tikz{
		\draw[-,black] (-5,0) edge[->,thick] (3.5,0);
		\draw[-,dashed,black] (-2.5,0.3) edge[-,thick] (-2.5,-0.25);
		\draw[-,dashed,black] (0.5,0.3) edge[-,thick] (0.5,-0.25);
		\draw[-,dashed,black] (-2.5,0.3) edge[-,thick] (0.5,0.3);
		\draw[-,dashed,black] (0.7,0.3) edge[-,thick] (0.5,-0);
		\draw[-,dashed,black] (0.7,0.3) edge[-,thick] (3.5,0.3);
		\draw[-,dashed,black] (-2.7,0.3) edge[-,thick] (-2.5,-0);
		\draw[-,dashed,black] (-5,0.3) edge[-,thick] (-2.7,0.3);
		\node[fill=white]  (v1) at (3,-0.5) {$\alpha\in\R$};
		\node[fill=white]  (v2) at (0.5,-0.5) {{$\rho(\wre)^{-1}$}};
		\node[fill=white]  (v3) at (-3.8,0.7) {{Sufficient}};
		\node[fill=white]  (v3) at (2.1,0.7) {{Can be unsatisfied}};
		\node[fill=white]  (v3) at (-1,1.15) {{Theoretical gap}};
		\node[fill=white]  (v3) at (-1,0.7) {{ITUC}};
		\node[fill=white]  (v2) at (-2.5,-0.5) {{$\eta(\wre)^{-1}$}};
		}
\caption{
\label{Schema} Interval of the Theoretical Unknown Conditions (ITUC). Theoretical gap between the scaling factor bounds for the necessary and sufficient condition of the Echo State Property. When $\alpha<\eta(\wre)^{-1}$ then the ESP is satisfied, when $\alpha>\eta(\wre)^{-1}$ then we can affirm that the ESP is satisfied.
}
\end{center}
\end{figure}

The stability also has been analyzed in~\cite{Yildiz2012}, the authors analyze a new sufficient and softer condition for the ESP. The ESP is studied in terms of the diagonal Schur stability, based on a positive definite matrix~\cite{Yildiz2012}.
As far as we know, there is a theoretical gap about the ESP existence when $\alpha$ belongs to the interval $U=[\eta(\wre)^{-1},\rho(\wre)^{-1}]$. When the scaling factor $\alpha$ belongs to $U$ the conditions about the ESN stability are unknown.
Figure~\ref{Schema} represents the theoretical results about ESP.
In~\cite{Zhang12} has been analyzed the asymptotic behaviour of this theoretical gap according characteristics of $\wre$. 
The authors using random matrix theory have proven that the size of  $U$ is large. 
The bound of the necessary condition is about twice the bound of the sufficient condition when the reservoir is composed by a very large pool of neurons (when $\Ns\rightarrow\infty$).
During this article we refer many times to the interval $U$, for this reason we  name $U$ as the Interval of the Theoretical Unknown Conditions (ITUC).
In this article, we study the accuracy of the model for reservoirs generated with $\alpha\in U$, when $U$ is an ITUC. On other words, we analyze the behaviour of models with scaled reservoirs with scaling factors in ITUC.

\section{Empirical evaluations}
\subsection{Methodology}
We analyze the behaviour of the canonical ESN when a random reservoir is generated with a scaling factor $\alpha$ in $U$, where $U$ is an ITUC defined in the previous section. 
We evaluate the accuracy of the model with the NRMSE on a group of well-known benchmark dataset. 
The problems are described in the next subsection.
As usual, we split the sequential data in two sets, one for setting the readout weights and another one is for their validation.
The error is computed applying free-run prediction (one step ahead). Them, the precedent predicted values are used as input patterns for predicting the next output. 
We define a grid of values for the reservoir size and the scaling factors.
This grid depends on the benchmark problem. Although, we always consider 10 different values of $\Ns$ and 10 different values of scaling factors $\alpha$.
In order of producing statistically significant results, we perform the experiments on a  benchmark dataset using $30$ different random initialisations.
For each specific benchmark problem, we arbitrary define $10$ reservoir size values $\Ns^{(1)},\ldots,\Ns^{(10)}$. For each reservoir size $\Ns^{(i)}$, we randomly initialize a reservoir matrix ${\wre}^{(i)}_{initial}$. 
Next, we compute the ITUC $U^{(i)}$. 
%
%
For each interval $U^{(i)}$, we compute $10$ values of scaling factors $\alpha^{(i,1)},\alpha^{(i,2)}, \ldots, \alpha^{(i,10)}$.
Then, we evaluate the model with a scaled reservoir matrix ${\wre}^{(i,j)}$, which is the original ${\wre}^{(i)}_{inital}$ after of being scaled with $\alpha^{(i,j)}$.
Note that, we repeat this experiment 30 times, therefore for each trial the interval $U^{(i)}$ is a different one, then the scaling factors are also different ones.

\begin{figure}[h]
 \begin{center}
   \includegraphics[width=0.8\linewidth]{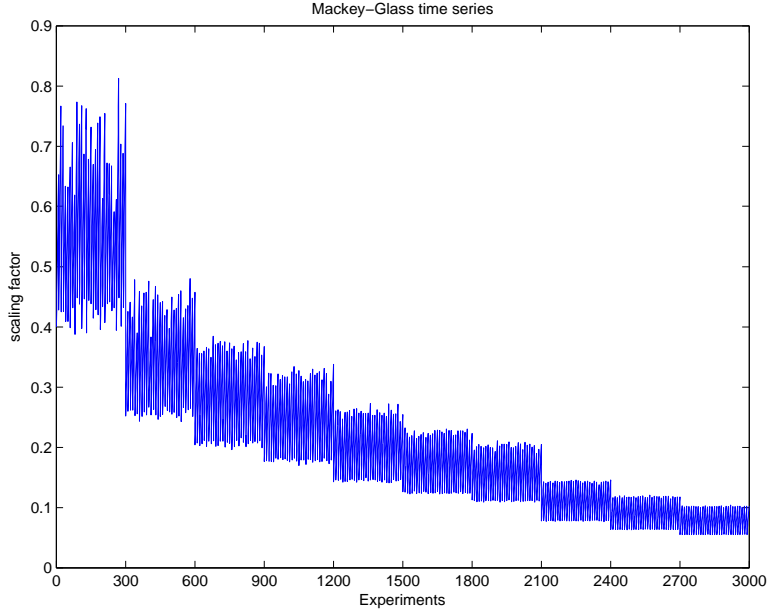}
\end{center}
\caption{\label{AlphaMackeyGlass} Scaling factor values applied for the Mackey-Glass benchmark problem. The first 300 experiments corresponds to a reservoir with 20 neurons, the next 300 experiments corresponds to a reservoir with 50 neurons, and so on. Larger experiment identification, then larger number of reservoir neurons is.}
\end{figure}

Figure~\ref{AlphaMackeyGlass} shows the different values of $\alpha$ when the problem was Mackey-Glass dataset. On the vertical axis are the scaling factor values and on the horizontal axis there are the experimental trials. The number of experimental trials for each benchmark problem was 3000 (total = number of repetitions (30) $\times$ different reservoirs (10) $\times$ different scaling factors (10)). The experiment number (experiment identification) increases with the larger of the reservoir, it means that the first 300 experiments corresponds to the smallest pool of reservoir, then the scaling factor is decreasing when the reservoir size is increasing. This is due to the fact that larger reservoir matrices have larger spectral radius and larger singular values~\cite{Zhang12}.

The input and reservoir weights are randomly initialised in the range $[-0.5,0.5]$. 
In this article we are using full connected reservoirs. 
Most often in the literature, a reservoir is built as a sparse pool of interconnected neurons (around $20\%$ of non zero values). 
However, there is an empirical evidence that the density of the reservoir matrix isn't a relevant factor on the model accuracy with respect to the relevance of the reservoir size and the spectral radius~\cite{MantasPracticalGuide12}.
In general, it is used sparse reservoirs only for computational reasons, because models with sparse matrices are faster than the models with dense ones.
All the simulations have been done in Matlab.

For each input pattern $\va\in\mathcal{A}$, the reservoir creates a high dimensional vector $\vs\in\mathcal{S}$. 
The dimension of  $\mathcal{S}$ is much larger than the dimension of $\mathcal{A}$.
There are several techniques for dimensionality reduction and visualization of high dimensional datasets. For example, these techniques include Metric Multidimensional Dimensionality Scaling~\cite{Sun2012}, PCA, Self-Organizing Maps~\cite{Kohonen96}, Sammon projections, Scale Invariant Maps~\cite{Fyfe05}, etc.
In order of analyzing how different values of $\alpha\in U$ can generate different reservoir projections, we define a multidimensional metric inspired of the techniques for dimensionality reduction mentioned above.
We define a metric that is a slight modification of the multidimensional scaling (MDS). Let $L(i,j)$ be the distance between two patterns $\va(i)$ and $\va(j)$ in the input space $\mathcal{A}$. Let $D(i,j)$ be the distance of two vectors on the projected space, that is the distance between $\vs(i)$ and $\vs(j)$ (the reservoir states generated by the network when the inputs are $\va(i)$ and $\va(j)$). In all the cases we are considering the euclidean distance. Then, we define the mean of the multidimensional scaling distance (we use the acronym MMDS), as follows:
\begin{equation}
\label{MMDS}
MMDS=\ds{\frac{1}{|\Delta t|}\sum_{\Delta t,i\neq j} \frac{(L(i,j)-D(i,j))^2}{D(i,j)}},
\end{equation}
where $\Delta t$ is some arbitrary range of time and we denote by $|\Delta t|$ the number of input patterns considered in this time range. Note that the form of MMDS is similar also to the Sammon error~\cite{Fyfe05}.
The goal of defining this measure is to have a notion about the topographic characteristic of the projections.
Small MMDS values are produced when $L(i,j)$ is near to $D(i,j)$.
On the other hand, large MMDS values are produced when close input patterns are projected far from each other.
\subsection{Benchmark Problems Description}
We analyze the reservoir projections using the following well-known simulated datasets:
\subsubsection{Mackey-Glass time-series}
Classic benchmark problem that has been analyzed in several papers on the RC area~\cite{Jaeger01,Gallicchio2011,Jaeger04}. The dynamics are given by:
$$
\ds{\frac{\partial u(t)}{\partial t}=\frac{0.2u(t-\tau)}{1+u(t-\tau)^{10}}-0.1u(t)},
$$
a common value for the parameter $\tau$ is $17$, due to the fact that when $\tau>16.8$ the system has a chaotic attractor~\cite{Gallicchio2011}.

\subsubsection{Noisy Multiple Superimposed Oscillator (MSO) time-series} 
The noisy MSO is a sequential dataset generated for two sine waves and gaussian noise.
The series is~\cite{Sheng2013}:
$$
a(t)=sin(0.2t)+sin(0.311t)+z, t=1,2\ldots,
$$
where $z$ is a Gaussian random variable with distribution $\mathcal{N}(0,0.01)$.
We simulate 10000 samples for training the model, and we present the performance of the trained model on 1000 unseen simulated samples.
\subsubsection{Lorenz attractor} 
The series is based on the Lorenz equations:
$$
\ds{
\frac{\partial x}{\partial t}=\sigma(y-x),
\frac{\partial y}{\partial t}=rx-y-xz, 
\frac{\partial z}{\partial t}=xy-bz,
 }
$$
we used the parameters $r=28$, $b=8/3$ and $\sigma=10$ and step size $0.01$. For more information about the integration of the ordinary differential equations is possible to see Runge-Kutta method~\cite{numericalRecips92}.
The training set has 13107 samples and the testing set contains 3277 samples.
Once the dynamics are simulated we normalize the data in the range [0,1].
\subsubsection{Rossler attractor} 
Classic time-series with a sequence generated for the dynamics: 
$$
\ds{
\frac{\partial x}{\partial t}=-z-y,
\frac{\partial y}{\partial t}=x+ry, 
\frac{\partial z}{\partial t}=b+z(x-c),
 }
$$
where the parameters values are $r=0.15$, $b=0.20$, $c=10.0$.
\subsubsection{Henon map} 
The Henon map is a well-known invertible mapping of a two-dimensional plane into itself~\cite{Henon76}.
The sequence is generated by: 
$$
x(t+1)=1-r x^{2}(t)+y(t),\quad y(t+1)=bx(t).
$$
where $r=1.4$, $b=0.3$ and initial states are $x=1$, and $y=1$.
Equivalent the sequence can be expressed as a 2-step recurrence as
$$
x(t+1)=1-r x^{2}(t)+bx(t-1).
$$
This sequence has been analysed with ESN in at least the following works~\cite{BasterCord11,Rodan11}.

\subsection{Empirical Results}
\begin{figure}[h]
 \begin{center}
   \includegraphics[width=0.99\linewidth]{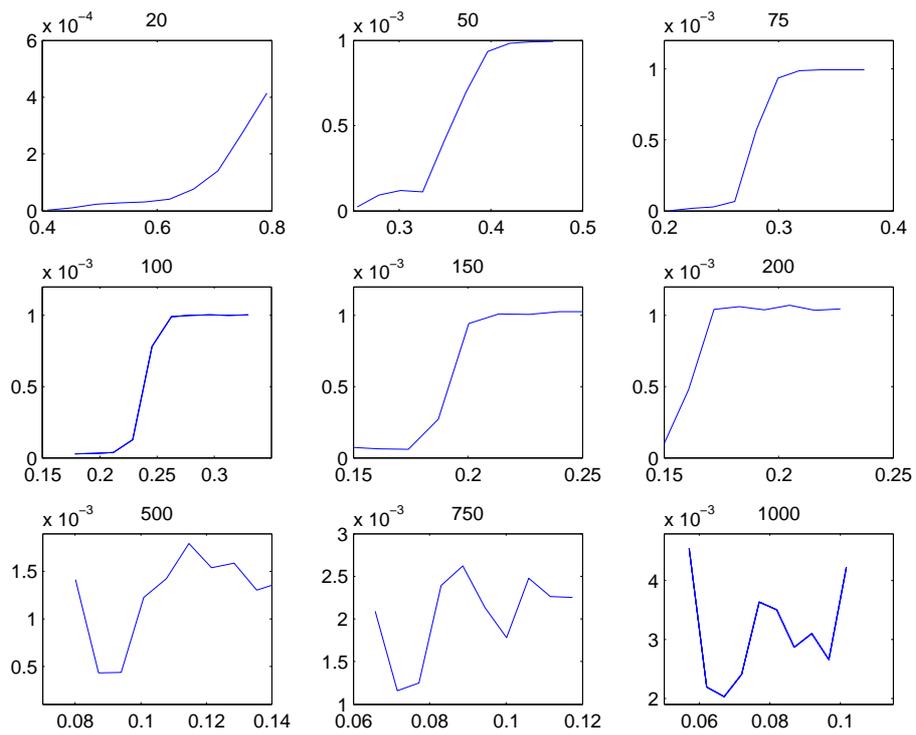}
\end{center}
\caption{\label{SubplotsMackeyGlass} Mackey-Glass dataset. Each subplot corresponds to o different reservoir sizes. The horizontal axis corresponds to the scaling factor, and the vertical axis corresponds to the NRMSE.}
\end{figure}

On the first benchmark problem we used a regularization factor on the ridge regression of $0.0001$, and reservoir sizes: $20$, $50$, $75$, $100$, $150$, $200$, $250$, $500$, $750$ and $1000$. On the rest of the problems the regularization factor was $0.001$ and reservoirs in $\{20,50,75,100,150,200,250,300,400,500\}$.
Figure~\ref{SubplotsMackeyGlass} shows several plots obtained with the Mackey-Glass time-series. 
Each one corresponds to a specific reservoir size, which is specified in the top of each graphic. 
The horizontal axis of each subplot corresponds to the scaling factor $\alpha$, and the vertical axis corresponds to the NRMSE.
Note that $U$ is different for each reservoir size. We can see that for the ``small'' reservoirs ($\Ns<150$), the accuracy is better when $\alpha$ is closer to the lower bound of $U$. On the other hand, for very large reservoirs the relationship between the accuracy and the scaling factor isn't clear.

\begin{figure}[h]
 \begin{center}
   \includegraphics[width=1\linewidth]{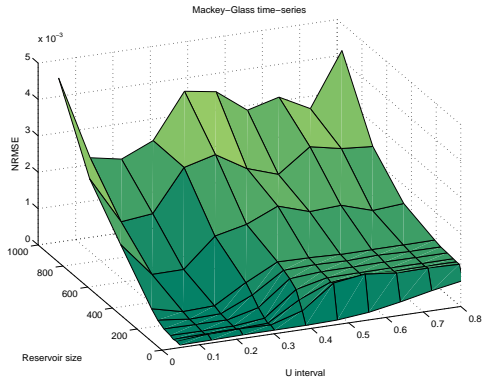}
\end{center}
\caption{\label{MackeyGlass} Mackey-Glass dataset. Accuracy with respect the scaling factor in $U$ and the reservoir size.}
\end{figure}

\begin{figure}[h]
 \begin{center}
   \includegraphics[width=1\linewidth]{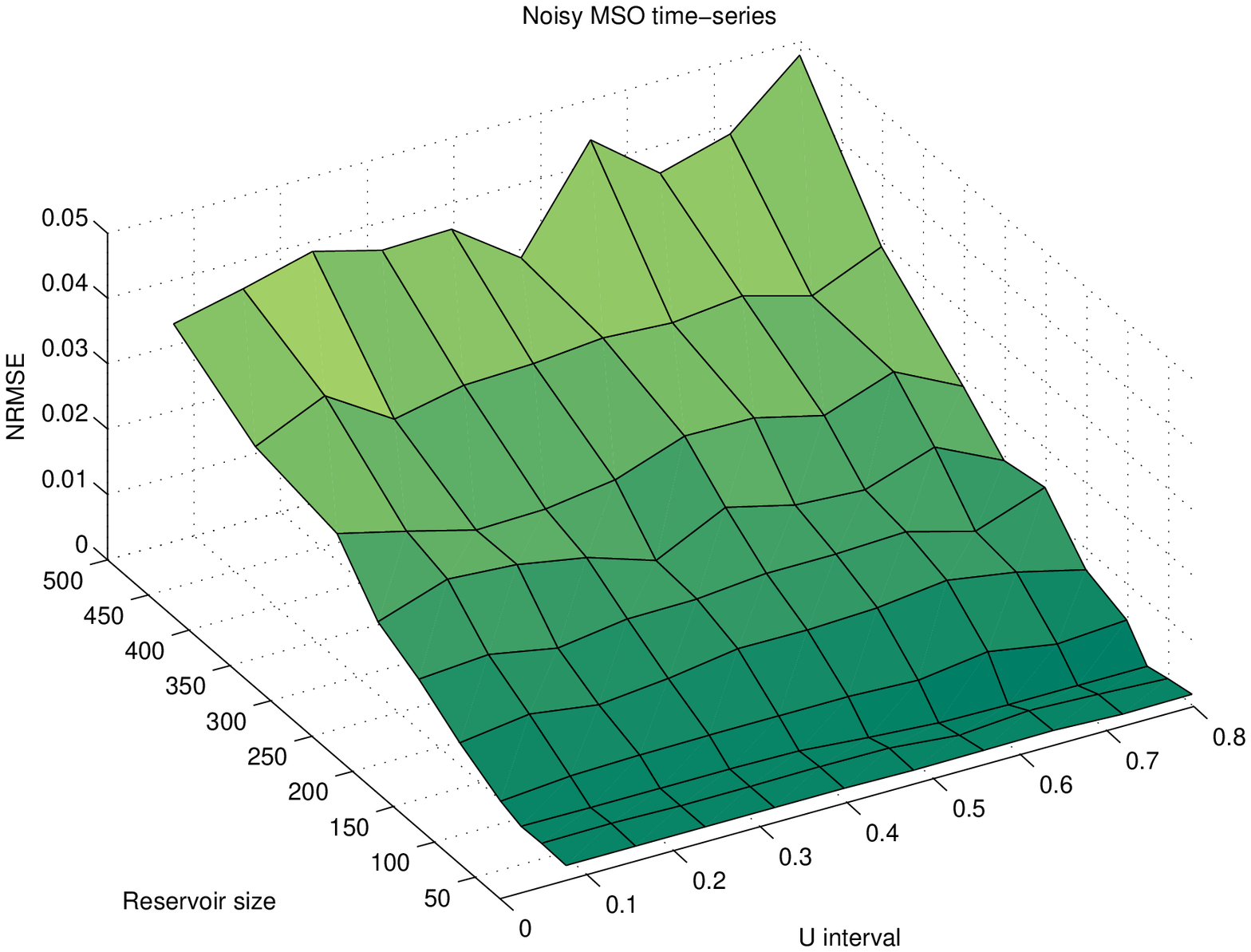}
\end{center}
\caption{\label{MSOSurf}  MSO dataset. Accuracy with respect the scaling factor in $U$ and the reservoir size.}
\end{figure}

\begin{figure}[h]
 \begin{center}
   \includegraphics[width=1\linewidth]{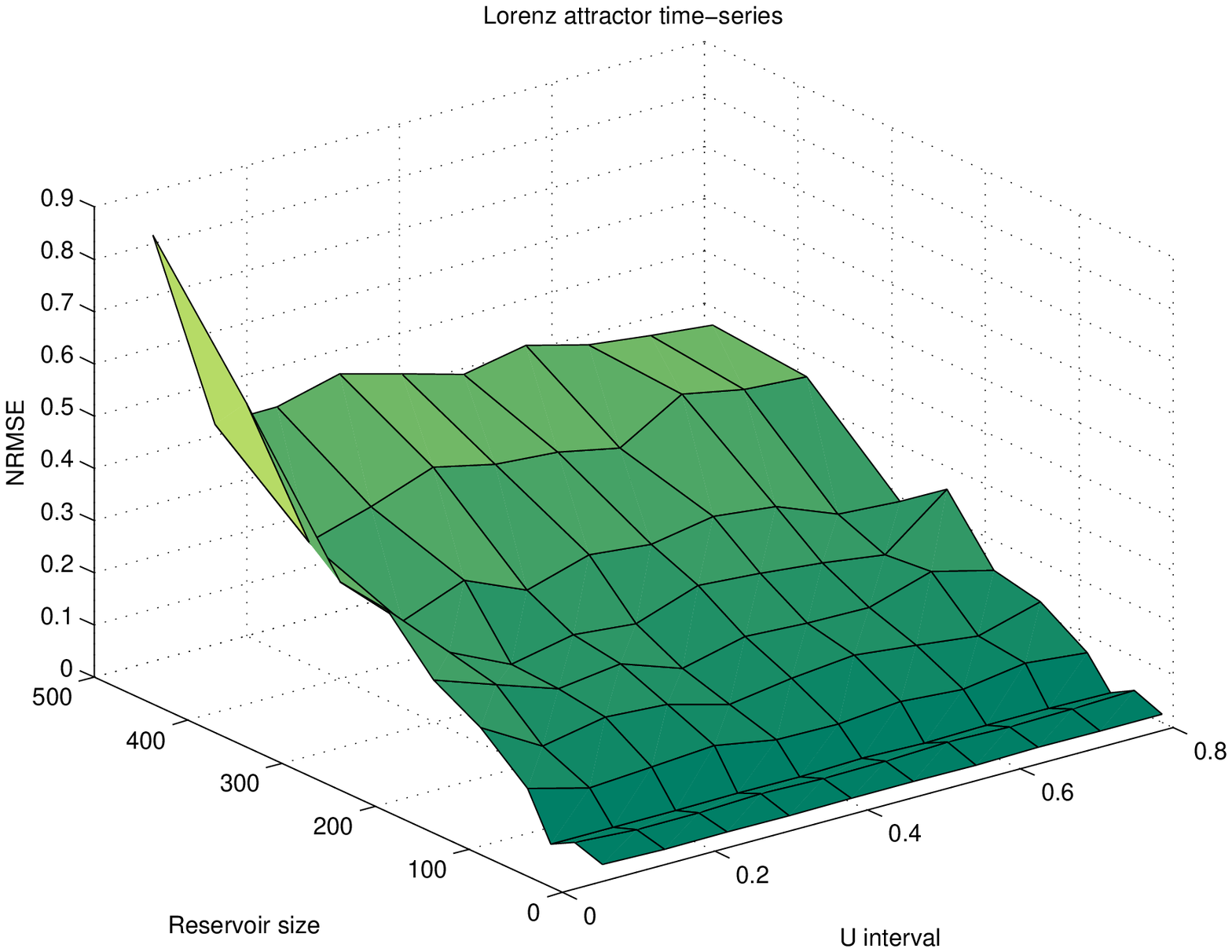}
\end{center}
\caption{\label{LorenzSurf} Lorenz dataset. Accuracy with respect the scaling factor in $U$ and the reservoir size.}
\end{figure}

\begin{figure}[h]
 \begin{center}
   \includegraphics[width=1\linewidth]{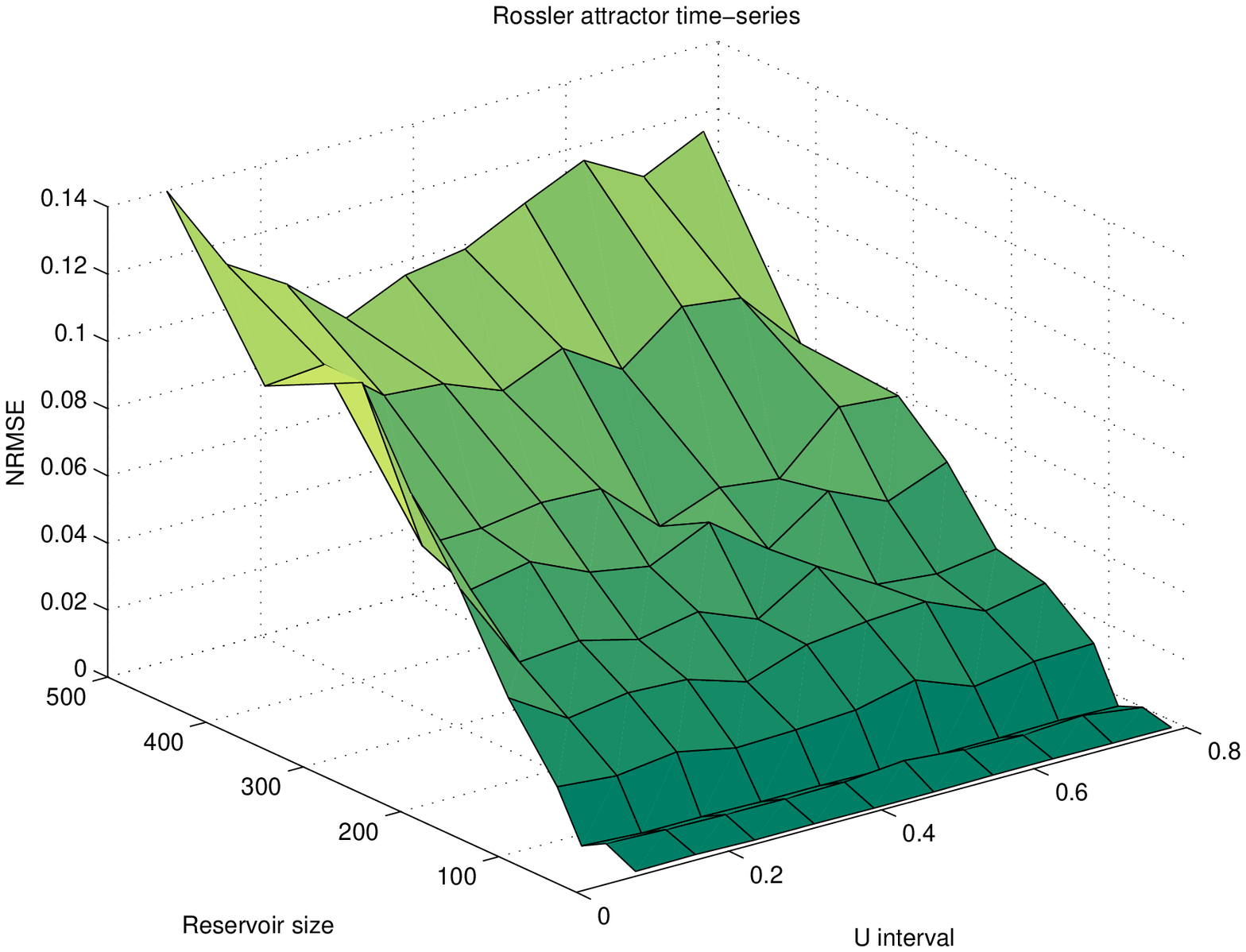}
\end{center}
\caption{\label{RosslerSurf}  Rossler dataset. Accuracy with respect the scaling factor in $U$ and the reservoir size.}
\end{figure}

\begin{figure}[h]
 \begin{center}
   \includegraphics[width=1\linewidth]{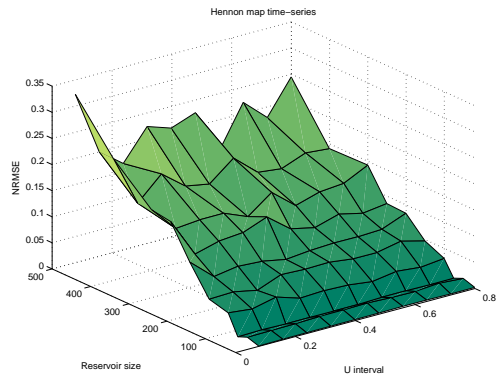}
\end{center}
\caption{\label{HenonSurf} Henon dataset. Accuracy with respect the scaling factor in $U$ and the reservoir size.}
\end{figure}

For each benchmark problems we are presenting two types of figures. 
One presents the accuracy NMSE with respect of the reservoir size and the $U$ interval. 
The another one presents the MMDS according to the reservoir size and the $U$ interval.
As we mentioned above, the $U$ interval depends of the reservoir size and the random initialization of the reservoir. Therefore, these graphics have been built as follows: for a specific reservoir size, we compute the $U$ interval, and a regular grid with 10 values. Then, we compute the average among the accuracy obtained on the 30 experiments.
Figures~\ref{SubplotsMackeyGlass} and~\ref{MackeyGlass} show that the scaling factor and the accuracy are sensible to the reservoir size. Extremely large reservoirs can be more unstable.
Figure~\ref{MSOSurf} shows (in the case of MSO dataset) that very large reservoirs and $\alpha$ values close to the upper bound of $U$ can cause unstable model accuracy.
When the reservoir is small, it seems that the behaviour of the reservoir projection is independent of the value of $\alpha\in U$.
Figure~\ref{LorenzSurf} shows the results for the Lorenz attractor benchmark problem. We can again see that small reservoirs are more stable, and it seems that the value of $\alpha$ doesn't impact on the accuracy when $\Ns$ is less than $200$.
On the other hand, for Lorenz attractor dataset the experiments with smaller $\alpha$ values and large reservoirs have the worst accuracy.
Figures~\ref{RosslerSurf} and~\ref{HenonSurf} show the accuracy obtained with Rossler attractor and Henon map datasets. Both figures have the same characteristics, the value of $\alpha$ seems to be less important on the accuracy than the reservoir size.

\begin{figure}[h]
 \begin{center}
   \includegraphics[width=1\linewidth]{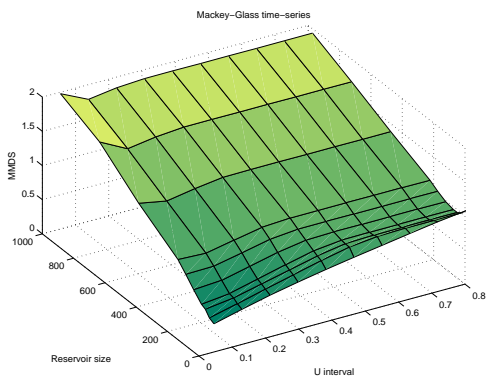}
\end{center}
\caption{\label{MackeyGlassMDS}  Mackey-Glass dataset. MMDS with respect the scaling factor in $U$ and the reservoir size.}
\end{figure}

\begin{figure}[h]
 \begin{center}
   \includegraphics[width=1\linewidth]{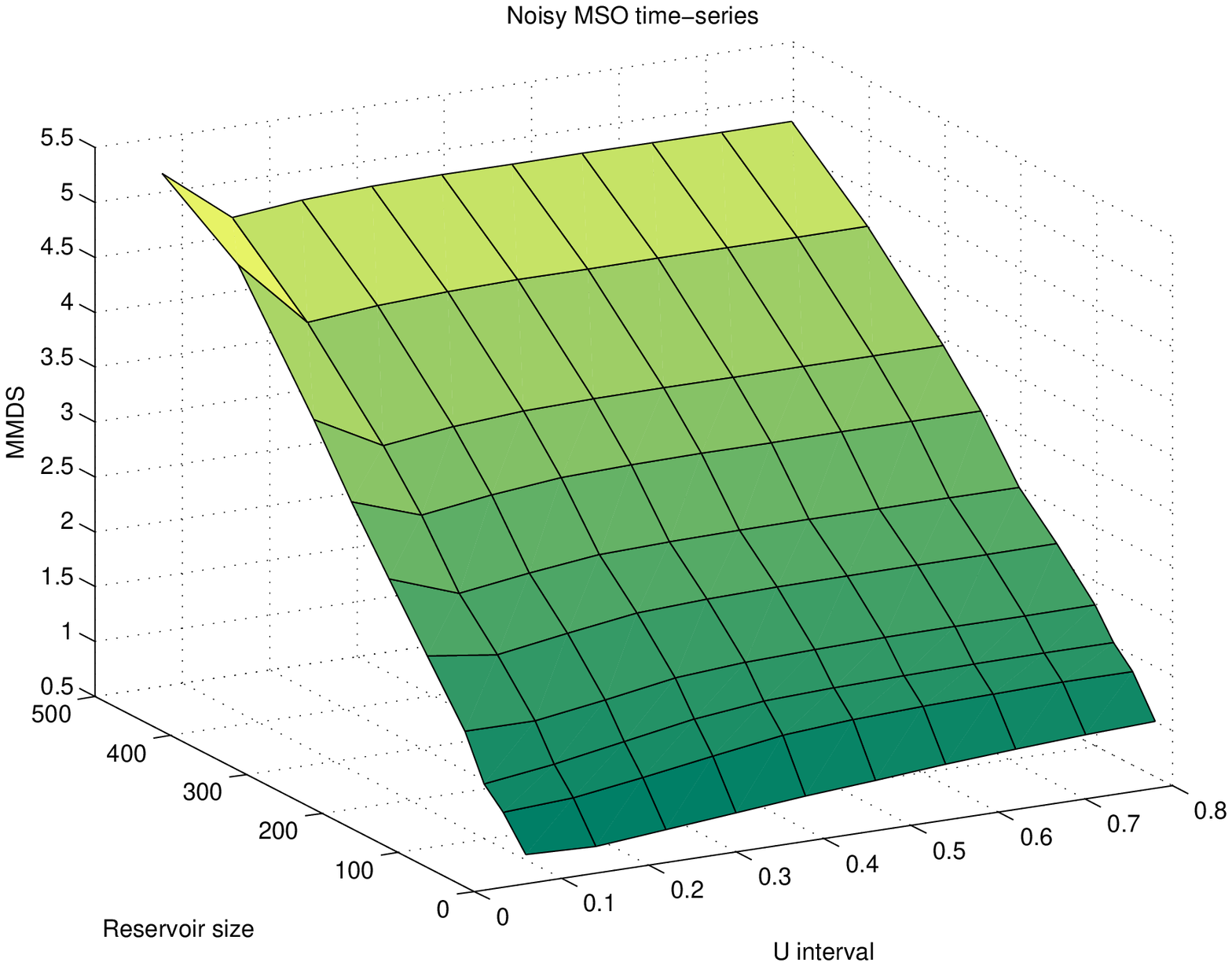}
\end{center}
\caption{\label{MSOMDS} MSO dataset. MMDS with respect the scaling factor in $U$ and the reservoir size.}
\end{figure}

\begin{figure}[h]
 \begin{center}
   \includegraphics[width=1\linewidth]{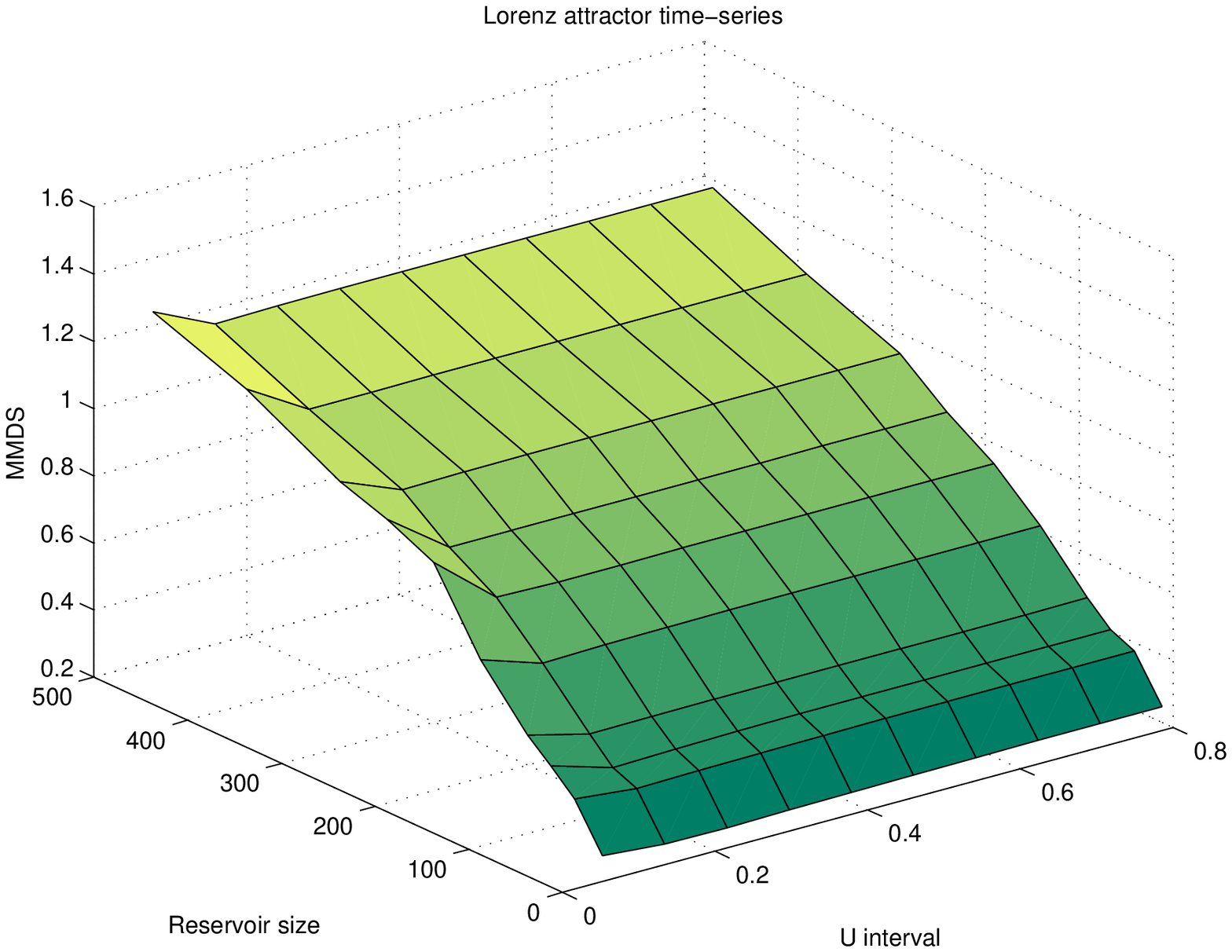}
\end{center}
\caption{\label{LorenzMDS} Lorenz dataset. MMDS with respect the scaling factor in $U$ and the reservoir size.}
\end{figure}

\begin{figure}[h]
 \begin{center}
   \includegraphics[width=1\linewidth]{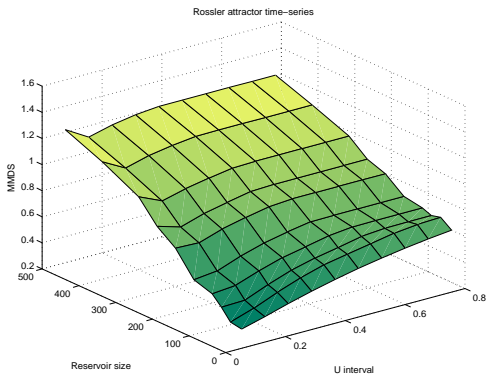}
\end{center}
\caption{\label{RosslerMDS} Rossler dataset. MMDS with respect the scaling factor in $U$ and the reservoir size.}
\end{figure}

\begin{figure}[h]
 \begin{center}
   \includegraphics[width=1\linewidth]{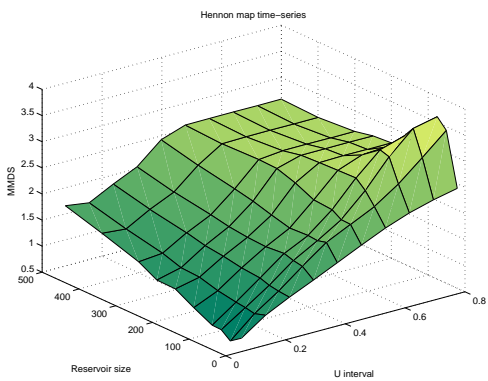}
\end{center}
\caption{\label{HenonMDS} Henon dataset. MMDS with respect the scaling factor in $U$ and the reservoir size.}
\end{figure}

Another group of pictures analyze how the scaling factor impact on the topographic characteristic of the reservoir projections. 
In general, we can see that larger reservoirs provoke larger MMDS values. 
However, the relationship between the MMDS values and $\alpha$ values depends on the benchmark data.
Figures~\ref{MSOMDS} and \ref{LorenzMDS} show how the MMDS is almost constant along the $U$ interval. 
On these figures the MMDS increases with the reservoir size.
The value of $\alpha$ seems to impact on MMDS measure according to the Figures~\ref{MackeyGlassMDS} and~\ref{RosslerMDS}. The impact seems to be less relevant than the impact of the reservoir size, but anyway we can see how larger values of $\alpha$ may cause larger values of MMDS.
A different behaviour occurs with the Henon map dataset, in Figure~\ref{HenonMDS} we can see that both the scaling factor and the reservoir size are relevant parameters.
A final remark, note that in almost all the benchmark problems the best accuracy occurs when the values of $\alpha$ are near to the lower bound of $U$.
As well as, in many cases the accuracy is stable for the different values of $\alpha$ in $U$.

\twocolumn


\onecolumn
\section{Conclusions}
%
%
A fundamental property of the Echo State Network (ESN) model is the Echo State Property (ESP), which impacts on the model predictions.
A sufficient condition for the ESP involves the singular values of reservoir matrix. On the other hand, a necessary condition for the ESP also has been introduced, the ESP is violated according to the spectral radius value of the reservoir matrix.
There is a theoretical gap between the necessary and sufficient conditions for the ESP.
We specify this gap in an interval named Interval of the Theoretical Unknown Conditions (ITUC), which is defined as function of the spectral and singular value of the reservoir matrix.
There is a large group of reservoirs, which we can't affirm that the ESP is satisfied nor ESP violation.
This article presents an empirical analysis of the accuracy and the projections of reservoirs that belong to this group.
According our experimental results, in some benchmark problems the best accuracies occur when the reservoirs are near to satisfy the sufficient condition for the ESP.
However, for small reservoirs with different spectral radius and singular values the accuracy obtained is stable. 
From previous works, is known that the optimal accuracy is obtained near to the border of stability control of the dynamics.
According to our results, it seems that this control border is closer  to the sufficient condition than to the necessary condition.
In addition, we studied the reservoir projections using a type of multidimensional scaling metric. We found different behaviour according to the benchmark problem.

In the near future, it can be interesting to analyze the ITUC using other metrics on the reservoir projections. For example, the exponential Lyapunov of reservoir projections created with scaling factor values in the ITUC.
In addition, the memory capacity when the scaling factor belongs to the ITUC can be also of interest for the community.

\clearpage



\bibliographystyle{plain}
\bibliography{refRnn}

\end{document}